\documentclass[10pt,twocolumn,letterpaper]{article}

\usepackage{iccv}
\usepackage{times}
\usepackage{epsfig}
\usepackage{graphicx}
\usepackage{amsmath}
\usepackage{amssymb}
\usepackage[ruled]{algorithm}
\usepackage{algpseudocode}
\usepackage{bbm}
\usepackage{verbatim}

\DeclareMathOperator*{\argmax}{arg\,max}
\newcommand{\p}{\,\mathrm{p}}
\newcommand{\ud}{\,\mathrm{d}}
\newcommand{\R}{\mathbb{R}}

\usepackage[pagebackref=true,breaklinks=true,letterpaper=true,colorlinks,bookmarks=false]{hyperref}

\iccvfinalcopy 

\def\iccvPaperID{2679} 

\ificcvfinal\fi

\begin{document}

\title{Minimum Delay Object Detection from Video}

\author{Dong Lao \,\, and \,\, Ganesh Sundaramoorthi\\
King Abdullah University of Science and Technology (KAUST), Saudi Arabia\\
{\tt\small \{dong.lao, ganesh.sundaramoorthi\}@kaust.edu.sa}
}

\maketitle
\ificcvfinal\thispagestyle{empty}\fi

\begin{abstract}
  We consider the problem of detecting objects, as they come into view, from videos in an online fashion. We provide the first real-time solution that is guaranteed to minimize the delay, i.e., the time between when the object comes in view and the declared detection time, subject to acceptable levels of detection accuracy. The method leverages modern CNN-based object detectors that operate on a single frame, to aggregate detection results over frames to provide reliable detection at a rate, specified by the user, in guaranteed minimal delay. To do this, we formulate the problem as a Quickest Detection problem, which provides the aforementioned guarantees. We derive our algorithms from this theory. We show in experiments, that with an overhead of just 50 fps, we can increase the number of correct detections and decrease the overall computational cost compared to running a modern single-frame detector.\footnote{Code: \url{https://github.com/donglao/mindelay}}
\end{abstract}

\section{Introduction}
Real-time closed loop systems continuously acquire data, process it, make decisions, and act to achieve some objective. An example includes a self-driving car. In a self-driving car, data is acquired, processed to make a decision on steering direction, and this decision is input to a control system that steers the car to achieve an objective such as avoiding pedestrians. In these types of closed loop systems, the data must be processed \emph{online}, i.e., as it is acquired. The decisions must be \emph{reliable} and must be made with an acceptable level of delay. For instance, in a self-driving car, a pedestrian must be detected reliably with little delay otherwise the decision to steer away could be too late for the control system to actuate and avoid a collision. Motivated by closed-loop system applications that acquire and process visual data, we are interested in developing computer vision algorithms that operate \emph{online}, and perform within limits on \emph{delay} and \emph{accuracy}.

In this paper, we look into a specific instance of this general problem in the context of object detection from video. We are interested in the problem of detection in closed loop scenarios. As the video is acquired, we want to as soon as possible 1) determine when an object of interest comes into view, and 2) we want to localize and determine the identity of the object at the frame the object comes into view. Further, we seek to operate under constraints on errors of detection, delay and computational cost. While deep learning has provided a wealth of object detectors \cite{ren2015faster,dai2016r,liu2016ssd,redmon2016you,lin2017focal} that operate on a single image and localize objects of interest, which in some cases are real-time, in many cases, they produce false alarms or fail to fire on objects due to phenomena such as partial occlusion, illumination, and other nuisances. Thus, although they may satisfy delay requirements, the detection accuracy may be poor. Of course one may leverage results over frames from a single-frame detector, i.e., several detections near the same location over multiple frames, ensures reliability. Because of this, there have been many works \cite{feichtenhofer2017detect,kang2017object,kang2018t,Liu_2018_CVPR,Zhu_2018_CVPR,Bertasius_2018_ECCV} that leverage temporal information over video batches to reduce false alarms. However, this comes at added delay before a detection can be declared and increased computational cost. Such delay may not be acceptable. Thus, in one case, one achieves acceptable delay but not detection accuracy, and in the other case one may achieve acceptable accuracy but not delay. In fact any algorithm will have to trade-off one for another.

In this paper, we design an algorithm for detection from videos that, for any given level of a false alarm constraint, minimizes the delay in detection. To do this, we leverage the \emph{Quickest Detection} theory \cite{poor2009quickest,veeravalli2013quickest} from the statistics literature. Quickest Detection addresses the problem of detecting changes in a stochastic process. It is assumed that the stochastic process is determined from some known probability distribution before some unknown change time, after which the stochastic process is determined from a different known distribution. The theory provides a means to derive an online algorithm to determine the unknown change time with \emph{minimum delay} subject to constraints on the false alarm rate or the minimum error subject to constraints on the delay. We pose our problem in that framework, leverage existing state-of-the-art single-frame detectors, and derive algorithms for guaranteed reliable object detection with minimum delay that operates in \emph{real-time}.

\subsection{Contributions}
Our specific contributions are as follows: {\bf 1}.~To the best of our knowledge, we introduce the first online, real-time, video object detector that guarantees minimum detection delay subject to given constraints on detection accuracy. {\bf 2}.~To do this, we formulate the minimum delay video object detector as a Quickest detection problem, and derive algorithms. {\bf 3}.~We provide a recursive approximation to the optimal algorithm, which empirically is shown to have similar detection performance as the optimal algorithm but operates in real-time. {\bf 4}.~We show with 50 fps overhead (un-optimized Matlab code) that we obtain more correct detections with less delay than single-frame detectors. We also show that the overall computational cost to achieve our detections is lower than single-frame detectors under the same levels of accuracy. {\bf 5}.~We introduce a performance analysis of online video object detectors that take into consideration both speed and accuracy, based on QD theory. This can be used to evaluate existing single-frame object detectors in the context of video applications.

\subsection{Related Work}
\textbf{Single Frame Object Detection}: Our work leverages methods for object detection from a single image. These methods take as input a single image and return bounding boxes localizing possible objects of interest; they also return class probabilities of the bounding box corresponding to an object class. Early works (e.g., \cite{dalal2005histograms,viola2001rapid}) for this problem use a sliding window approach along with a classifier trained with traditional machine learning. Currently, CNN based approaches are the dominant approach. There are two families of such detectors: 1) \emph{two-stage detectors} (e.g., \cite{ren2015faster,dai2016r}) which generate region proposals for likely locations of objects, then solve the classification problem via a CNN for each proposed bounding box, and 2) \emph{one-stage} detectors (e.g., \cite{liu2016ssd,redmon2016you,lin2017focal}) which predict the bounding boxes and their class information in one step. The latter are often computationally less costly, but may be less accurate than the former \cite{huang2017speed}. As we will show in this paper, when video is available, all these detectors can be significantly improved in terms of computational time before which an object is detected at any level of detection accuracy.

\textbf{Video-based Data Association}: There is a substantial literature, sometimes referred to as \emph{data association} (e.g., \cite{zhang2008global,huang2008robust,okuma2004boosted}), which relates to a sub-task of the problem we consider in this paper. In the data association problem, given a batch of frames from a video and the output of a single-frame object detector on each of those frames, the goal is to associate or link the bounding boxes corresponding to the same object across frames to produce trajectories. This can then be used in a number of applications, such as object tracking and action recognition. Recent works, e.g., \cite{feichtenhofer2017detect, Chen_2018_CVPR}, make use of deep learning to determine the links and refine them along with the detections in a joint fashion. Similar to this literature is work on determining \emph{tublets}, similar to trajectories, from video motivated by the Imagenet-VID challenge \cite{russakovsky2015imagenet}. These works (e.g., \cite{kang2017object,kang2018t}) make use of CNNs to predict spatio-temporal volumes corresponding to an object over frames, and then an LSTM (a recurrent neural network) to classify the object.

These methods can be used for detection of objects in video to provide more temporal consistent results, though adapting them recursively and real-time is not straightforward. Further, these methods do not address the issue of how small the batch size could be chosen to guarantee an acceptable detection accuracy. Larger batches lead to more reliable detections, but with larger delay and computational cost. Our work explicitly addresses the trade-off between delay (computational cost) and detection accuracy, and provides a guaranteed minimum delay solution.

{\bf Online Object Tracking}: The literature on online object tracking is extensive, and we do not intend to give a review. In this literature, one is given an initial bounding box of the object, and the goal is to determine it in subsequent frames in an online fashion. For instance, \cite{breitenstein2009robust,bolme2010visual,Ma_2015_CVPR,danelljan2015convolutional} use correlation filters for tracking, and recent works (e.g., \cite{wojke2017simple,bae2018confidence}) apply deep learning. These works do not address of problem of \emph{detection}, as that is explicitly assumed in the first frame; one may use our method to initialize such trackers.
 
\textbf{Online Detection in Videos}: Our work relates to \cite{lao2017minimum}, which addresses the online detection of moving objects from video using motion cues. There, a minimum delay solution with given accuracy constraints is formulated. However the method is far from real-time due to expensive invocations of optical flow and a non-recursive algorithm.  In this paper, we leverage existing CNN-based single-frame detectors rather than motion and derive a recursive solution to provide a \emph{real-time} solution. Another method that examines the trade-off between speed and accuracy is \cite{chen2015speed}, motivated by biological systems. A related online method to our work is \cite{Shou_2018_ECCV}, which is a method for determining the start of an action. This method, however, does not address issues of delay versus accuracy.

\section{Review of Quickest Detection Theory}
We briefly highlight the main concepts in Quickest Detection, and refer the reader to \cite{veeravalli2013quickest} for a more detailed survey. Consider a stochastic process $\{X_t \}_{t=0}^{\infty}$. Before an unknown {\it change time} $\Gamma$, $X_t$ has distribution $\p_0$ and after the change time $\Gamma$, $X_t$ has distribution $\p_1$. Quickest Detection (QD) aims at reliable \emph{online} determination of distributional changes with minimum delay, i.e., minimum time after the change time. The main idea is that reliability can be obtained by observing more (noisy) data, but with added delay, and the theory seeks to provide algorithms addressing this trade-off.

In QD, a {\it stopping time} $\tau$ is a function of the data $\{X_t\}_{t=0}^s$, i.e., realizations of the stochastic process, up to the current time $s$ that returns $s$ when it declares a change has occurred at some time before $s$. QD seeks to find an optimal stopping time, with respect to the optimization problem defined next. The {\it average detection delay} of $\tau$ is
\begin{equation}
  \mathsf{ADD}(\tau) = \sup_{t\geq 1} \mathbb E_t [ \tau - t | \tau \geq t  ]
\end{equation}
where $\mathbb E_t$ is the expectation given the change time is $t$. This defines the worst case average delay over all change times. The {\it false alarm rate} is defined as $\mathsf{FAR}(\tau) = 1 / \mathbb{E}_{\infty} [ \tau ]$, that is, one over the average stopping time given that there is no change. QD solves the following optimization problem:
\begin{equation} \label{eq:stoppingrule_opt}
  \min_{\tau} \mathsf{ADD}(\tau) \mbox{ subject to } \mathsf{FAR}(\tau) \leq \alpha,
\end{equation}
where $\alpha \in [0,1]$ is the maximum tolerable false alarm rate. This formulation
recognizes that without any constraints on the false alarm rate, that the optimal stopping rule is simply to declare a detection in the first frame, which would yield a large false alarm rate, and thus QD imposes a constraint.

It can be proven that the optimal stopping rule that solves the above optimization problem is obtained by computing the following {\it likelihood ratio}: 
\begin{equation} \label{eq:max_likelihood}
  \Lambda_t = \frac{ \mathbb{ P } [ \Gamma < t | X_1, \ldots, X_t ] }
  { \mathbb{ P } [ \Gamma \geq t | X_1, \ldots, X_t ] }
  = \max_{1\leq t_c < t } \prod_{s=t_c}^t  \frac{ p_1(X_s) } {  p_0(X_s) },
\end{equation}
where the second equality is with the additional assumption that $X_i$ are iid. The optimal stopping rule then declares a change at the first time $t$ that the likelihood exceeds a threshold, i.e., $\Lambda_t>T(\alpha)$. The threshold $T$ is a function of the $\mathsf{FAR}$ constraint, and the distributions.

The aforementioned test in the iid case has a recursive implementation, thus the maximization need not be explicitly computed, as follows:
\begin{align}
    &\tau = \inf\{n\geq1: W_n\geq \log T(\alpha)\}\\
      &W_t =\left[\max\limits_{1\leq t_c\leq t}\sum_{i=t_c}^t\log\frac{\p_1(X_i)}{\p_0(X_i)}\right]^+\\
      &W_{t+1} = [W_t + \log\p_1(X_{t+1})-\log\p_0(X_{t+1})]^+ \label{eq:cusum_original}
\end{align}
where $[\bullet]^+\triangleq \max[\bullet,0]$. One declares a change when $W_t$ exceeds a threshold. This recursive likelihood ratio test is named the {\it cumulative sum} (CUSUM) \cite{page1954continuous} algorithm.

In many applications, like ours, the distributions may not be fully known, and may depend on an unknown parameter $\theta$. In this case, one can estimate and re-estimate the parameter $\theta$ via a ML or MAP estimation at each time $t$, and still guarantee optimality of the test in \eqref{eq:max_likelihood}. This, however, does not extend itself to a recursive implementation.~\cite{siegmund1995using} and \cite{lai1998information} provide different methods in this scenario.

\section{Minimum Delay Object Detector}

In this section, we formulate our minimum delay object detector that operates on video by the use of Quickest Detection theory. We first introduce the problem setting and notation, and then proceed to deriving detection algorithms.

\subsection{Notation and Assumptions}\label{notations}
We denote by $b = (x, y, \ell_x, \ell_y)\in \R^4$ a \emph{bounding box} (of an object in an image) where $(x,y)$ is the centroid and $\ell_x$ and $\ell_y$ are the $x$- and $y$-scales of the bounding box. We denote $B \subset \R^4$ to be the space of all bounding boxes in the image under consideration. A \emph{trajectory} is a sequence of bounding boxes over consecutive frames; this will be denoted as $b_{t_s,t_e} \triangleq (b_{t_s}, b_{t_s+1}, \ldots, b_{t_e})$ where $t_s$ and $t_e$ are start and end times and $b_t$ denotes a bounding box at time $t$.

An image from a video sequence at time $t$ will be denoted $I_t$. A \emph{single-frame object detector} operates on an image and outputs a collection of bounding boxes, which we denote $B_{obs} \subset B$ and call the \emph{observed bounding boxes}, of possible locations of objects in an image. It also outputs the probabilities that each bounding box $b\in B_{obs}$ corresponds to one of $n+1$ classes of semantic object categories. These categories are denoted $l_0, \dots, l_n$ where $l_0$ corresponds to the ``background" or the class of objects not of interest. The class probabilities for a particular bounding box $b\in B_{obs}$ are denoted $v_i(b) \triangleq \p(l = l_i|I_t, b)$, and the vector of all such probabilities over all classes is denoted $v(b) \triangleq (v_0(b),\dots, v_n(b))^T$. In addition, two-stage detectors, e.g., Fast-RCNN, output a confidence score $\mu(b)\in \R^+$ that the bounding box $b\in B_{obs}$ corresponds to an object.

For convenience in later computations, we will use the function, which we call the \emph{data at time $t$}, $D_t : B \to [0,1]^{n+1} \times B$ that maps a bounding box in image $I_t$ to class probabilities and the bounding box itself, i.e., $D_t(b)\triangleq (v(b), b)$. If $A \subset B$, we define $D_t(A) \triangleq \cup_{b\in A} D_t(b)$. For a given image, the output of the function $D_t$ will only be known for the observed bounding boxes, $B_{obs}$. We will see that it will be important in our algorithm to also estimate probabilities involving $D_t$ even in the set of bounding boxes for which the detector does not output class probabilities. Therefore, we introduce nomenclature for this set, called the \emph{unobserved bounding boxes}, defined by $B_{unobs} \triangleq B \backslash B_{obs}$.

We let $I_{t_s,t_e} \triangleq (I_{t_s},I_{t_s+1}, \ldots,I_{t_e}$) and $D_{t_s,t_e} \triangleq  (D_{t_s}, D_{t_s+1}, \ldots, D_{t_e})$, where $t_e$ and $t_s$ are start and end times.

\subsection{Formulating the Object Detector from QD}
\begin{figure*}[t]
\begin{center}
\includegraphics[width=\linewidth]{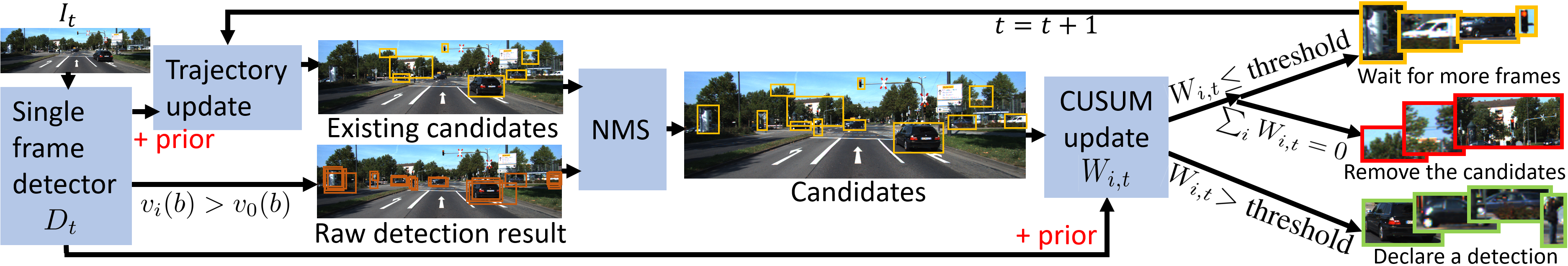}
\end{center}
   \caption{{\bf Schematic of Our Minimum Delay Detector.} The output from single-frame CNN detectors are data input to our method. The additional computation (at roughly 50 fps) performed recursively provides reliable detection results with minimum delay. See Algorithm~\ref{alg:detection}.}
   \label{fig:schematic}
\end{figure*}

We are interested in detecting objects of interest, i.e., only those belonging to the pre-specified categories $l_1, \ldots, l_n$, in the scene as soon as they come into view of the observer. To do so, we setup a Quickest Detection problem for \emph{each} object of interest in the scene. Each object is characterized by its trajectory $b_{1,t}$ from the start time $1$ to the current time $t$, which indicates the object's projection into the imaging plane. Given an estimate of this trajectory, which we estimate and update sequentially, we wish to determine if the object of interest is in view of the observer at time $t$, by posing this as a hypothesis testing problem. The null hypothesis is that the trajectory $b_{1,t}$ describes bounding boxes that do not correspond to a consistent object of interest (i.e., the trajectory corresponds to regions in the images of class $l_0$) and the alternative hypothesis is that the object remains out of view (or consists of bounding boxes that are of class $l_0$) up until a time $\Gamma_{0,i}$, which we call the \emph{change time}, at which point the object is in view and thus the bounding boxes $b_{\Gamma_{0,i},t}$ correspond to a class $l_i$.

At each time $t$, the data available from which we may make the decision of object in view is denoted $D_{1,t}$, which is the output of the single-frame detector at each frame from $1$ to $t$, consisting of class probabilities in the observed set $B_{obs}$ as well as the unobserved set $B_{unobs}$ with class probabilities unknown. Even though one does not have direct measurements of class probabilities in the latter set, we assume prior class probabilities. According to QD, we estimate $\p(\Gamma_{0,i}<t|D_{1,t}, b_{1,t})$, which is the probability that the object is in view before time $t$ given the data up to time $t$ and conditioned on a trajectory $b_{1,t}$ that must be estimated. The trajectory is analogous to a parameter $\theta$ in QD of the distributions that is unknown. We also estimate $\p(\Gamma_{0,i}\geq t|D_{1,t}, b_{1,t})$, i.e., the probability that the object is not in view before or at time $t$.

According to QD, the optimal detection rule is a threshold of the likelihood ratio, $\Lambda_t$, i.e., the max over all object classes $l_i$ of the two aforementioned probabilities:

\begin{align}\label{eq:lr_test}
    \Lambda_{t}(b_{1,t})&=\max\limits_i\frac{\p(\Gamma_{0,i}<t|D_{1,t},b_{1,t})}{\p(\Gamma_{0,i}\geq t|D_{1,t},b_{1,t})}\\
    &= \max\limits_i\max\limits_{t_c \geq 1}\frac{\p_i(D_{t_c,t}|b_{t_c,t})}{\p_0(D_{t_c,t}|b_{t_c,t})}
\end{align}
where for simplicity of notation, we set $\p_i(\bullet)\triangleq\p(\bullet|l=l_i)$. Note the data $D_t$ across frames along the known trajectory $b_{1,t}$ of object class $i$ is independent across frames. This is because knowing the object identity along the trajectory removes the class information from the data, which results in random nuisances that are assumed independent. This statement remains true for data near the trajectory due to spatial regularity of the single-frame detector. As our algorithm will only consider data near the trajectory, we assume this is true for all data. This gives that 
\begin{align}
    \Lambda_{t}(b_{1,t}) &=\max\limits_i\max\limits_{t_c \geq 1}\prod_{j = t_c}^{t}\frac{\p_i(D_j|b_j)}{\p_0(D_j|b_j)}.\label{eq:lr_simple}
\end{align}

A detection is declared when $\Lambda_{t} > T(\alpha)$. Here $T(\alpha)$ is a threshold chosen according to a given false alarm constraint $\alpha$. The detected object class is the $i^{\ast}$ for which the maximum of \eqref{eq:lr_simple} over $i$ is achieved.

\subsection{Estimating the Trajectory}\label{sec:data_association}

In the previous sub-section, it was assumed that the trajectory $b_{t_c,t}$ of the object of interest was given, however, it is unknown and must be estimated from the data. We now discuss its estimation via an optimization problem, and divulge the solution to a later section. As stated earlier, the trajectory is treated analogously to an unknown parameter $\theta$ of the pre- or post-change distribution in the QD problem. We may estimate that parameter in a number of ways, including a maximum-likelihood or MAP estimation if a prior probability is known. Estimation in these ways guarantees optimality of the detection rule. Since we wish to incorporate a smoothness prior, we use a MAP estimator.

Our prior assumption on the trajectory $b_{1,t}$ is that it is class-independent and of nearly constant velocity in each of the parameters of the bounding box. Therefore, we consider
\begin{equation}\label{eq:constant_speed}
    \log \p(b_{t_1,t_2}) \propto -\sum_{t=t_1+1}^{t_2-1}\lVert b_{t-1} - 2b_{t} + b_{t+1}\rVert_2^2.
\end{equation}
This prior ensures that paths that are nearly linear have higher prior probability than other paths. One may easily adapt any other assumption about the trajectory accordingly. For example, \cite{xing2009multi,bae2014robust, park2015minimum} provide a series of different techniques for this task. The MAP estimator for the trajectory $b_{1,t}$ given the data $D_{1,t}$ is then as follows:
\begin{align}
b^*_{t_c,t} &= \argmax\limits_{b_{1,t}}\max\limits_i \p(b_{1,t})\prod_{j = t_c}^{t}\p_i(D_j|b_j),\label{eq:b_est}
\end{align}
which is just the numerator in the likelihood ratio multiplied by the prior. Note that for each candidate change time $t_c$, one has to estimate the trajectory from above. However, in Section~\ref{sec:simplification}, we show how this can be avoided for efficiency. In the next sub-section, we describe how to simplify the likelihood terms so that we can then solve this estimation problem as well as determine the full likelihood ratio.

\subsection{Computing Pre- and Post-Change Probabilities}
In order to compute the likelihood ratio in \eqref{eq:lr_simple} as well as in the estimation of the trajectory \eqref{eq:b_est}, one needs to compute $\p_i(D_t|b_t)$. To evaluate this probability, we separate the data $D_t$ into the data from the observed set $B_{obs}$, and the un-observed set $B_{unobs}$. Therefore,
\begin{align}\label{eq:observed_decompose}
\p_i(D_t|b_t) & = \p_i(D_t(B_{obs})\cup D_t(B_{unobs})|b_t) \nonumber\\
&= \int_{B_{obs}\cup B_{unobs}} \p_i(D_t(b)|b_t) \ud \mu(b)
\end{align}
where $\mu(b)$ is the measure of bounding box $b$. For $b\in B_{obs}$, we set $\mu(b)$ equal to the confidence score from the Region Proposal Network when using two-stage detectors, while $\mu(b)$ is constant when using one-stage detectors. For the unobserved part, $\mu(b)$ is also assumed to be constant.

\indent {\bf Computing the Probability, $\p_i(D_t(b)|b_t)$}: We simplify $\p_i(D_t(b)|b_t) = \p_i(v(b), b|b_t)$ by noting that $v(b)$ and $b$ are independent, i.e., not knowing the image, the output of class probabilities from a single-frame detector is independent of location, as they are built invariant to location. Therefore,
\begin{align}
    \p_i(v(b), b|b_t)&= \p(v(b)|l=l_i)\p(b|b_t)\\
    &= \frac{\p(l = l_i|v(b))\p(v(b))}{\p(l = l_i)}\p(b|b_t)\\
    &\propto\frac{v_i(b)\p(b|b_t)}{\p(l = l_i)}\label{eq:detection_flip}
\end{align}
where we have used that $v_i(b) = \p(l = l_i|v(b))$, i.e., given all the class probabilities, the probability of class $l_i$ is just the $i$-th component of $v(b)$, and that $\p(v(b))$ is a constant due to positional invariance of the single-frame detector. $\p(l = l_i)$ is the prior probability of the object classes.

Following the loss function used in training single-frame object detectors, we set $\p(b|b_t)$, i.e., the probability of $b$ knowing the true location $b_t$, to be one if the intersection of union score between the bounding boxes, $\text{IoU}(b,b_t)$, surpasses a fixed threshold and zero otherwise, i.e.,
\begin{equation}\label{eq:p(b|b)}
   \p(b|b_t) =\mathbbm{1}\{\text{IoU}(b,b_t) > IoU_{lim}\}.
\end{equation}

{\bf Computing the Probability, $\p_i(D_t(B_{unobs})|b_t)$}: Now we compute $\p_i(D_t(B_{unobs})|b_t)$ by setting the class probabilities of a bounding box to be the same as the class prior probabilities, i.e., $v_i(b) = \p(l=l_i)$, which in the absence of data is a reasonable assumption, and the confidence measure of a bounding box to be $\mu(b)=\textit{constant}$ for all unobserved $b$. Thus, we see that \eqref{eq:detection_flip} becomes
\begin{equation}\label{eq:unobserved_simplify}
\p_i(D_t(b)|b_t) \propto \frac{\p(l=l_i)}{\p(l=l_i)}\p(b|b_t) = \p(b|b_t), \quad b\in B_{unobs}.
\end{equation}
Therefore,
\begin{align}
\!\!\!\!\p_i(D_t(B_{unobs})|b_t) &\propto \int_B\p(b|b_t)\ud \mu(b) - \int_{B_{obs}}\p(b|b_t) \ud \mu(b)\nonumber\\
    &= C-\sum_{b \in B_{obs}}\p(b|b_t)\mu(b)\label{eq:unobserved_simplify}
\end{align}
where we treat $C=\int_B\p(b|b_t)\ud \mu(b)$ as a constant (independent of $b_t$) that is chosen so that the overall probability above is positive, and is set empirically as discussed below.

{\bf Computing the Probability, $\p_i(D_t|b_t)$}: Now we can compute the full probability $\p_i(D_t|b_t)$ by combining \eqref{eq:detection_flip}, \eqref{eq:p(b|b)} and \eqref{eq:unobserved_simplify}, which yields
\begin{align}
& \p_i(D_t|b_t)  \propto \int_{B_{obs}} \frac{v_i(b)\p(b|b_t)}{\p(l=l_i)} \ud \mu(b) + \p_i(D_t(B_{unobs})|b_t)\nonumber\\
&= \sum_{b \in B_{obs}} \left[ \frac{v_i(b)}{\p(l=l_i)}-1\right]\mathbbm{1}\{\text{IoU}(b,b_t) > IoU_{lim}\}\mu(b) + C.\label{eq:detector_computation}
\end{align}

Note that \eqref{eq:detector_computation} is computed by summing, over all observed bounding boxes that are close spatially (with respect to the IoU metric) to the given bounding box $b_t$, a measure of how informative of object class $i$ the single-frame detection of $b$ is over the prior of the detector weighted by the confidence $\mu(b)$ that the box is an object class of interest. The constant $C$ can be interpreted as a prior on the trajectory. Large values of $C$ favors greater dependence on the prior $p(b_{t_c,t})$ in the MAP estimation problem \eqref{eq:b_est} and so the estimated trajectory more likely follows a constant velocity path. This also means that the likelihood ratio accumulates more slowly, but is more robust to imperfections in the data such as failure due to partial occlusion, illumination, etc. Therefore, $C$ controls the robustness to the imperfections.

\subsection{Summing Up: Detection Algorithm}
Our algorithm for minimum delay object detection is described in three steps, which are iterated as new data $D_{t+1}$ becomes available, as follows: 1) update of the existing trajectories via the MAP estimation \eqref{eq:b_est}, 2) new trajectory generation, and 3) evaluation of the likelihood ratio $\Lambda_{t+1}$ test \eqref{eq:lr_simple}.  Algorithm~\ref{alg:detection_full} describes this process. We discuss the first two steps in more detail in the paragraphs below.

{\bf Trajectory Update}: At each time $t$, we have a set of candidate trajectories, $b_{1,t}^k, \, k = 1, \ldots, n_{traj}$. We wish to update them into frame $t+1$. At time $t+1$, the data $D_{t+1}$ from the single-frame detector is available. The update of the existing trajectories into frame $t+1$ is done by solving the MAP estimation problem \eqref{eq:b_est} for each of the existing trajectories. This is done by running iterative updates of each bounding box in the trajectory to maximize the objective alternatively. The process is initialized with the trajectory extended into frame $t+1$ with the constant velocity model. This optimization process also computes $\p_i(D_{t+1}|b_{1,t+1}^k)$.

Note in the version of Quickest Detection with parameter estimation, new data in future frames can impact the estimation of the unknown parameter, in our case the trajectory, and thus the change time $t_c$ and the likelihood ratio, which could lead to faster detection. However in our specific setup, additional locations on the trajectory predicted before the current estimate of $t_c$ would have already been initialized with our trajectory spawning scheme (below) at the time before $t_c$, therefore, we neglect re-estimating $t_c$, which saves considerable computational cost.

{\bf Trajectory Generation}: We now propose new candidate trajectories as follows. We use the data from the frame $t+1$ to determine candidate bounding boxes $b_{t+1}^{new,k}$ by choosing $b$ such that the object class $i$ probability is greater than background probability, $v_i(b) > v_0(b)$. We use those boxes and their class probabilities and perform non-max suppression with the bounding boxes $b_{t+1}^k$ and probabilities $\p_i(D_{t+1}|b_{t,t+1}^k)$ from existing trajectories. The change time $t_c$ of these newly spawned trajectories is $t+1$.

\begin{algorithm}[t]
  \begin{algorithmic}[1]
    \State{$t=0$}
    \State run single-frame detector on $I_{t}$ and obtain $D_t$.\label{step:start_update1}
    \State Find new candidates s.t. $v_i(b)>v_0(b)$ and apply NMS
    \For {each candidate}
    \State Update the trajectory by \eqref{eq:b_est}.
    \State Update likelihood ratio by computing \eqref{eq:lr_simple} for all $i$.
    \If{$\Lambda_{t}>\text{threshold}$,} declare a detection, output the position $b^*_t$ and label $l_i$
    \EndIf
    \EndFor 
    \State{$t = t+1$}\label{step:end_update1}. Repeat \ref{step:start_update1}-\ref{step:end_update1}.
  \end{algorithmic}
  \caption{\sl Minimum Delay Object Detection (Full)}
  \label{alg:detection_full}
\end{algorithm}

In the next section, we avoid the expensive requirement of updating the whole trajectory at each time $t$ and revisiting data $D_{1,t}$ by providing a recursive update of the trajectory to obtain a fully recursive algorithm. We also introduce further pruning of candidate trajectories. Although this recursive procedure does not theoretically guarantee optimality of the delay, we analyze the empirical performance against the optimal Algorithm~\ref{alg:detection_full} and show that little is lost in delay with considerable gains in speed.

\section{A Recursive Approximation for Speed}\label{sec:simplification}

We now present a recursive approximation of the trajectory computation, which allows us to derive a fully recursive algorithm allowing one to avoid re-visiting previous data from the single-frame detector.

\subsection{Recursive Trajectory / Likelihood Computation}
To estimate the trajectory $b_{t_c,t}$ recursively in the MAP estimation problem, we decompose the prior of the trajectory \eqref{eq:constant_speed} as follows:
\begin{equation}
         \p(b_{t_c,t}) = \p(b_{t_c})\prod_{k=t_c+1}^t \p(b_k|b_{t_c,k-1}).
\end{equation}
Instead of going back to all previous frames, we only estimate the bounding box $b^*_t$ at the current frame by assuming the trajectory at the previous frames are optimized. Therefore, we need only consider the term $\p(b_t|b_{t_c,t-1})$ as the prior in the MAP estimation problem. With the constant speed assumption, this term becomes
\begin{equation}
    \log \p(b_t|b_{t_c},b_{t-1},\dots) \propto -\lVert b_{t-2} - 2b_{t-1} + b_{t}\rVert_2^2.
\end{equation}
The MAP estimation problem in \eqref{eq:b_est} then becomes equivalent to solving 
\begin{align} 
b^*_t &=\argmax\limits_{b_{t}}\p_i(D_t|b_t)\p(b_t|b_{t_c,t-1})\\
&=\argmax\limits_{b_{t}}[\log\eqref{eq:detector_computation} -\lVert b_{t-2} - 2b_{t-1} + b_{t}\rVert_2^2]\label{eq:association_recursive},
\end{align}
as the terms $\p_{i}(D_{s}|b_s)$ for $s<t$ are independent of $b_t$.

With this approximation, we may compute the likelihood ratio $\Lambda_{t}(b_{1,t})$ in \eqref{eq:lr_simple} recursively by use of the CuSum algorithm \eqref{eq:cusum_original}. Defining $W_{i,t} = [\log{\Lambda_{t}}]^+$, the recursive update of $W_{i,t}$ is as follows:
\begin{align}
&W_{i,t_c} = 0,\nonumber\\
&W_{i,t} =[W_{t-1} + \log\p_i(D_t|b^*_t)-\log\p_0(D_t|b^*_t)]^+. \label{eq:recursive_update}
\end{align}
As soon as $W_{i,t}$ exceeds a threshold, determined by the false-alarm rate, a detection is declared.

Note that the non-recursive algorithm in the previous section has computational complexity $\mathcal{O}( n\times t^2 )$ where $n$ is the number of objects of interest and $t$ is the time the detection is declared, as at each time, we have to revisit the data up to the current time. The recursive implementation considered in this section has complexity $\mathcal{O}( n\times t )$, a considerable savings, which we explore further in experiments.

\subsection{Further Simplifications and Final Algorithm}
Our final simplified algorithm is  Algorithm~\ref{alg:detection} (see also Figure~\ref{fig:schematic}), which summarizes the recursive approximation described in the previous section, and involves two additional simplifications, described below.

\begin{algorithm}[t]
  \begin{algorithmic}[1]
  \State{$t=0$}
    \State run single-frame detector on $I_{t}$ and obtain $D_t$.\label{step:start_update}
    \State Find new candidates s.t. $v_i(b)>v_0(b)$ and apply NMS
    \For {each candidate}
    \State Predict the trajectory by \eqref{eq:association_recursive}.
    \State Update CUSUM statistic by \eqref{eq:recursive_update}.
    \If{$\sum_i W_{i,t} = 0$,} remove this candidate.
    \ElsIf{$W_{i,t}>\text{threshold}$,} declare a detection, output the position $b^*_t$ and label $l_i$
    \EndIf
    \EndFor
    \State{$t=t+1$}\label{step:end_update}. Repeat \ref{step:start_update}-\ref{step:end_update}.
  \end{algorithmic}
  \caption{\sl Recursive Minimum Delay Object Detection}
  \label{alg:detection}
\end{algorithm}

\textbf{Reduction of Class-Dependent Trajectories}:
When updating the trajectory by \eqref{eq:association_recursive}, one would have to find the best $b^*_t$ for each $i$. However, we only update the trajectory for the object classes satisfying $W_{i,t} > 0$. This is because if $W_{i,t}=0$, the likelihood ratio is less than 1, indicating the change time is in the future, eliminating the need for consideration of the trajectory under the class $i$ assumption.

\textbf{Removing Trajectories}:
We remove candidate trajectories if $\sum_i W_{i,t} = 0$, i.e., $W_{i,t}=0$ for all $i$. In this case, the trajectory does not carry any information about the object.

\section{Experiments}

\subsection{Datasets}
To test our algorithm, we need a video dataset that contains objects of multiple object classes, objects that appear at various unknown times, and all frames in each video are annotated. To the best of our knowledge, the best dataset that fits all these criteria is the KITTI dataset \cite{Geiger2012CVPR}.

This dataset contains 21 road scene videos and 917 annotated objects (tracklets) including cars, vans, bicycles and pedestrians. The dataset contains significant occlusion, illumination change, and viewpoint changes. Every visible object of these classes are annotated in every frame. Each object has an ID and becomes visible at an unknown frame. We set the ground truth change time of each object to be the first frame that it is annotated.

\subsection{Performance Metrics}
The output of a detection method is bounding boxes with class declarations and times, which represent when the method first detects objects. We make the following definitions for empirical quanitities. A \textbf{correct detection} is a detection whose bounding box overlaps with a ground truth bounding box in the same frame with IoU over $IoU_{lim}$ and the label matches the ground truth. Note an object may be detected multiple times, but this is only counted once for each ground truth object. A \textbf{false alarm} is a declared detection that is not a correct detection.

We use the following performance metrics. The \textbf{false alarm rate} is the number of false alarms divided by the total number of declared detections for the entire dataset. The \textbf{detection delay} is number of frames between when the object is detected minus the change time, i.e., the ground truth frame this object first appears. The \textbf{average detection delay} is the average of delay of all objects annotated in the dataset. If a ground truth object is not detected, it has maximum delay, which is the last frame it is annotated minus the ground truth change time.

\subsection{Methods Evaluated}
\textbf{Single-Frame Detectors:} We test our algorithm with both one-stage and two-stage detectors. We choose SSD \cite{liu2016ssd} and Retinanet \cite{lin2017focal} for one stage detectors, and the Faster-RCNN \cite{ren2015faster} (two-stage) as the single-frame detectors. For Faster-RCNN, we use the original implementation from the authors trained on Pascal-VOC07/12 as the baseline method. The backbone networks are  ZF \cite{zeiler2014visualizing} and VGG-16 \cite{simonyan2014very}, and the recent Resnet50 \cite{he2016deep}. For Resnet50 Faster-RCNN and SSD network, we use the implementation from \textit{mmdetection} \cite{mmdetection2018} toolbox.

\textbf{Comparisons:}
We use the direct detection results from single-frame detectors to compare to our method. Different false alarm rate levels are achieved by thresholding the detection response. Since single-frame detectors do not address temporal connection between frames, bounding boxes in adjacent frames are grouped into the same trajectory if the overlap is over $IoU_{lim}$. For each object, the detection delay is computed based on the first correct detection.

For our proposed method, for single-stage detectors that do not output a $\mu(b)$ from the RPN, we manually set $\mu(b) = 1$ for all observed bounding boxes. In all experiments we fix $IoU_{lim} = 0.5$. We set the prior probability $\p(l=l_i)$ of the object class to be uniform. The constant $C$ is set empirically for each single-frame detector.

\def\fWidFAR{\textwidth}
\begin{figure}[t]
\centering
\minipage{0.27\textwidth}
  \hspace*{-2mm}\includegraphics[width=\fWidFAR]{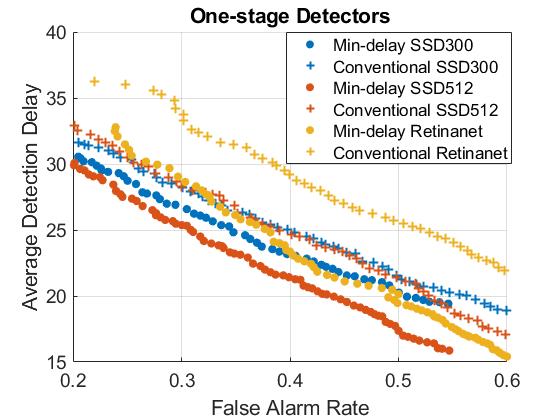}
\endminipage
\minipage{0.23\textwidth}
\hspace*{-4mm}\includegraphics[width=\fWidFAR]{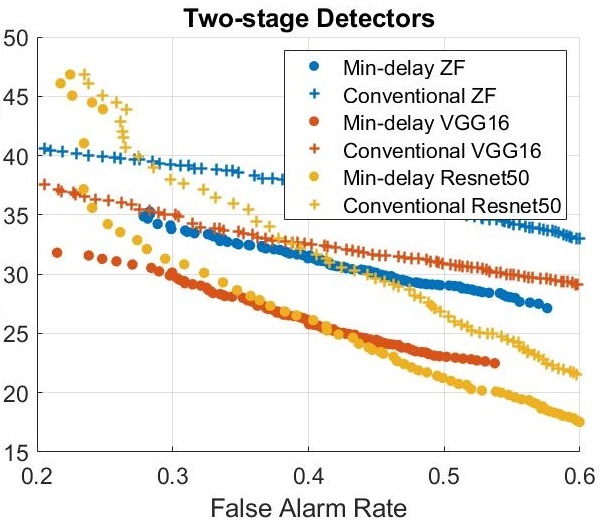}
\endminipage
\caption{{\bf Delay vs False Alarm Rate.} Compared to single-frame detectors, our method achieves less average delay at any FAR.}
\label{fig:far_delay}
\end{figure}

\subsection{Results}
\textbf{False Alarm Rate vs Delay}: Figure \ref{fig:far_delay} plots the false alarm rate verses delay curve by varying the detection threshold. Under all false alarm rates and every single-frame detector, our algorithm has less delay.

Interestingly, single-frame SSD300 and SSD512 have almost identical performance, however, the minimum delay version of SSD512 outperforms minimum delay SSD300. This indicates that SSD512 has more consistent detection results over frames compared to SSD300, thus allowing the likelihood to accumulate more quickly.

\begin{figure}[t]
\begin{center}
\minipage{0.272\textwidth}
   \hspace*{-2mm}\includegraphics[width=\fWidFAR]{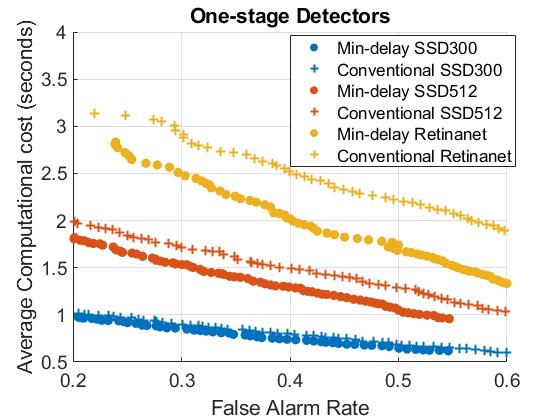}
 \endminipage
 \minipage{0.23\textwidth}
    \hspace*{-4mm}\includegraphics[width=\fWidFAR]{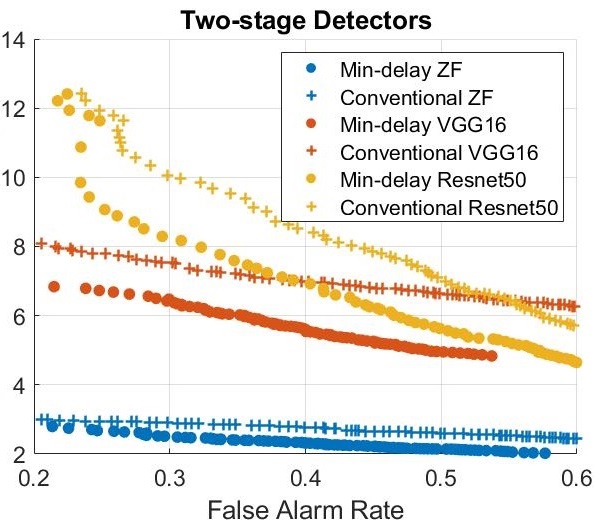}
  \endminipage
\end{center}
   \caption{{\bf Average Computational Time.} Our method achieves less computational cost than single-frame detectors. The result shows that noisier detectors (e.g. SSD300 and ZF) achieve less computational cost at any FAR by running over multiple frames at a faster speed than more-accurate detectors run on few frames.}
\label{fig:far_delay_cost}
\end{figure}

\textbf{Detection Accuracy vs Computational Cost}:
Figure \ref{fig:far_delay_cost} shows the average computational cost for detecting an object in seconds. In real-time online applications, the computational resources of the system are always limited. The result shows that one can use a faster but noisier single-frame detector, and still achieve lower overall computational cost under any accuracy constraint by using multiple frames.

\textbf{Analysis of Performance Gains}: Figure~\ref{fig:improvement} shows a more detailed analysis of the performance gains. Under all false alarm levels, the minimum delay detector outputs more correct detection results than the baseline, and these correct detections happen with lower delay.
\def\fWidFAR{\textwidth}
\begin{figure}[ht]
\begin{center}
\minipage{0.263\textwidth}
\hspace*{-3mm}\includegraphics[width=\fWidFAR]{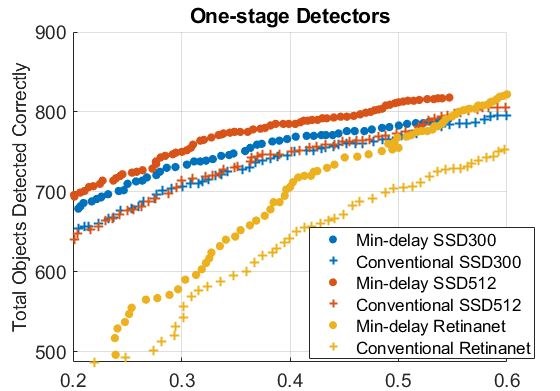}
 \endminipage
 \minipage{0.233\textwidth}
\hspace*{-4.7mm}\includegraphics[width=\fWidFAR]{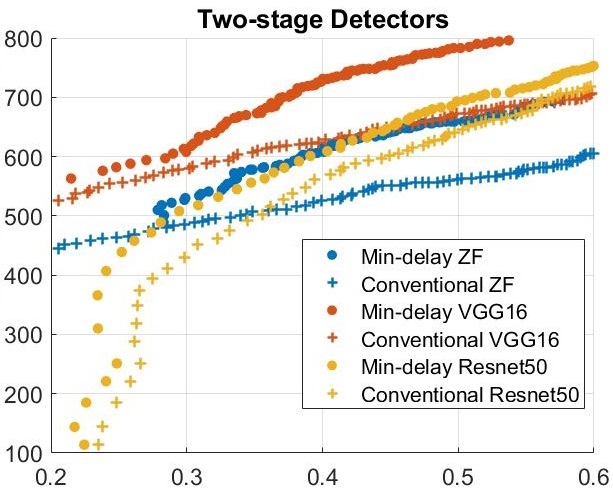}
 \endminipage\hfill
\minipage{0.275\textwidth}
\hspace*{-3mm}\includegraphics[width=\fWidFAR]{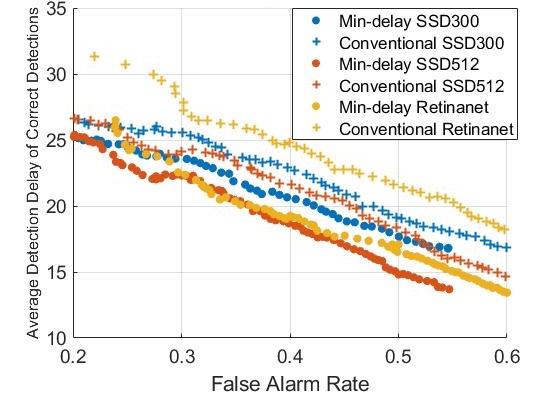}
 \endminipage
 \minipage{0.233\textwidth}
\hspace*{-6mm}\includegraphics[width=\fWidFAR]{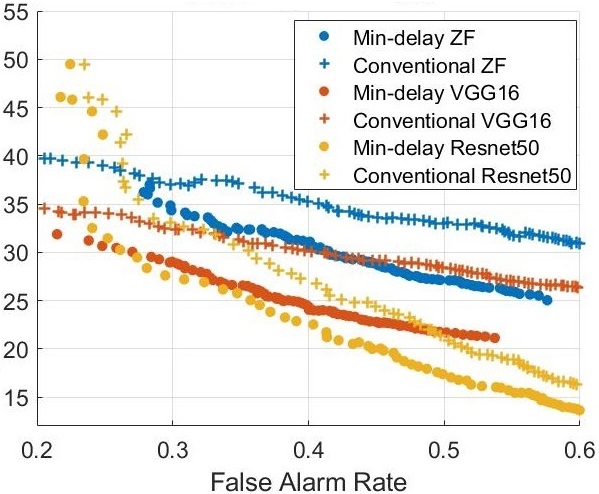}
 \endminipage
\end{center}
   \caption{{\bf Analysis of Performance Gains}: [Top]: Our method correctly detects more objects, and [Bottom]: with less average detection delay of correct detections than single-frame detectors.}
\label{fig:improvement}
\vspace{-1mm}
\end{figure} 

\begin{figure}[ht]
\begin{center}
\minipage{0.28\textwidth}
    \hspace*{-5mm}\includegraphics[width=\fWidFAR]{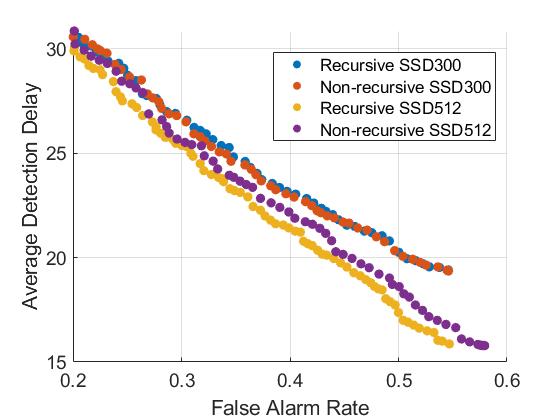}
     \endminipage
    \minipage{0.245\textwidth}
  \hspace*{-8mm}\includegraphics[width=\fWidFAR]{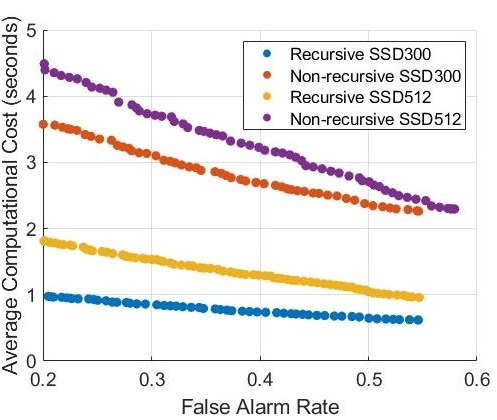}
   \endminipage
\end{center}
   \caption{{\bf Recursive vs Non-Recursive Algorithm.} The recursive approximation preserves the optimality of the algorithm while achieving significantly less computational cost.}
\label{fig:recursive}
\vspace{-2mm}
\end{figure}

\textbf{Recursive vs Non-Recursive Detection}: 
We compare the recursive approximation to the non-recursive version of our algorithm. We use SSD300 and SSD512 for illustration. Figure \ref{fig:recursive} shows the false alarm rate vs delay and computational cost curve. We find the result from the recursive version of the detector is comparable to the non-recursive counterpart while saving considerable computational cost. In SSD512, the recursive version reaches slightly better, though not significant (i.e., one-frame), performance than the non-recursive version.

\textbf{Computational Cost}:
On KITTI, our recursive algorithm typically runs at 40-100 fps with a Matlab implementation (excluding the cost of the single image detection process) depending on the number of objects visible in the scene. A single-frame detector such as SSD-300 runs in 59 fps, and thus our overall algorithm runs at 24-38 fps.

\section{Conclusion}
Our online object detector that operates on video achieves guaranteed minimum delay subject to false alarm constraints following theoretical results from QD. Further, our novel recursive formulation provided significant computational cost savings over the QD optimal detector and almost no loss in performance. Empirically, we showed that our recursive formulation achieves less delay and computational cost than single-frame detectors for any level of false alarm rate. Our method uses single frame detectors and uses \emph{simple} additional logic that runs in roughly 50 fps, and when combined with a single frame detector that is also real-time, results in a real-time algorithm. Thus, this has potential to be used in real-time closed loop applications. Additionally, our algorithm allows single image deep learning detectors to be applied to video without any additional training and have guaranteed minimum delay at any accuracy level.
{\small
\bibliographystyle{ieee_fullname}
\bibliography{detection}
}

\end{document}